\documentclass{article}

\usepackage{arxiv}

\usepackage[utf8]{inputenc} % allow utf-8 input
\usepackage[T1]{fontenc}    % use 8-bit T1 fonts
\usepackage{hyperref}       % hyperlinks
\usepackage{url}            % simple URL typesetting
\usepackage{booktabs}       % professional-quality tables
\usepackage{amsfonts}       % blackboard math symbols
\usepackage{nicefrac}       % compact symbols for 1/2, etc.
\usepackage{microtype}      % microtypography
\usepackage{lipsum}
\usepackage{subfigure}
\usepackage{graphicx}
\usepackage{algpseudocode,algorithm,algorithmicx}
\usepackage{wrapfig}
\usepackage{mathtools}
\usepackage{amsmath}
\usepackage{amsfonts}
\usepackage{amssymb}

\usepackage{flushend}

\title{Evolving Neural Networks in Reinforcement Learning by means of UMDA$_c$}

\author{
  Mikel Malagon \\
  University of the Basque Country (UPV/EHU)\\
  Donostia-San Sebastian, Spain\\
  \texttt{mmalagon002@ikasle.ehu.eus} \\
   \And
  Josu Ceberio \\
  University of the Basque Country (UPV/EHU)\\
  Donostia-San Sebastian, Spain\\
  \texttt{josu.ceberio@ehu.eus} \\
}

\begin{document}
\maketitle

\begin{abstract}
Neural networks are gaining popularity in the reinforcement learning field due to the vast number of successfully solved complex benchmark problems. In fact, artificial intelligence algorithms are, in some cases, able to overcome human professionals. 
Usually, neural networks have more than a couple of hidden layers, and thus, they involve a large quantity of parameters that need to be optimized.
Commonly, numeric approaches are used to optimize the inner parameters of neural networks, such as the stochastic gradient descent. However, these techniques tend to be computationally very expensive, and for some tasks, where effectiveness is crucial, high computational costs are not acceptable. 

Along these research lines, in this paper we propose to optimize the parameters of neural networks by means of estimation of distribution algorithms. More precisely, the univariate marginal distribution algorithm is used for evolving neural networks in various reinforcement learning tasks.
For the sake of validating our idea, we run the proposed algorithm % \textcolor{red}{(neural networks with estimation of distribution algorithm)} 
on four OpenAI Gym benchmark problems. In addition, the obtained results were compared with a standard genetic algorithm. 
Revealing, that optimizing with UMDA$_c$ provides better results than the genetic algorithm in most of the cases.
\end{abstract}

% keywords can be removed

\section{Introduction}

% NNak gaur egun 
% In the last decades, computers have experienced huge progress in terms of hardware. Nevertheless, every day problems that humans solve almost without effort, such as, natural language processing, face recognition, speech recognition, machine vision...  are in some cases very time consuming in terms of computational effort for many artificial intelligence algorithms.

% NN-en lorpenak
Recent advances in artificial intelligence have permitted computers to effectively deal with some real-world problems, such as natural language processing, face recognition, speech recognition, machine vision and more \cite{haykin1994neural}. Furthermore, in some of these tasks artificial intelligence has outperformed humans \cite{krizhevsky2012imagenet}. 
 %For example, DeepMind's AlphaGo was first AI to defeat a Go world champion in October 2015 \cite{chouard2016go}. As noted previously, this algorithms can also be used to solve real-world problems, in fact, DeepMind's AlphaFold is being used to predict 3D shapes of proteins based only in their genetic sequence.

% RL garrantzia 
In artificial intelligence and, more precisely, in machine learning, Reinforcement Learning (RL) is a field of great relevance. This is partially motivated by the large quantity of benchmark problems that are being solved in this field in recent years. RL algorithms are capable of developing complex behaviors from high dimensional input data, enabling algorithms to solve difficult problems such as playing 3D games \cite{lample2017playing}. 
% RL deskribapena
In RL, an agent takes actions in an environment and a reward value is given back. The agent has to learn an optimum behavior in order  maximize the cumulative reward. However, learning the optimum behavior is usually a hard optimization problem. 
In this sense, firstly the agent or the set of agents needs to be defined. There exist many ways of implementing the agents, for example as a vector of its own features, or, as is the case of agents controlled by NNs, as vectors of the parameters of NNs.
%There are many ways for implementing agents in RL tasks, to give some examples, defining agent's features in a vector or by means of NNs.
% NN zuzenketa bat...
In spite of the existence of several types of NNs, such as deep neural networks (DNNs) \cite{haykin1994neural}, recurrent (RNN) \cite{mikolov2010recurrent} and convolutional (CNN) \cite{krizhevsky2012imagenet}, in this paper feed-forward NNs (FNNs) are used due to their simplicity, Figure \ref{fig:feed-forward}.

In many cases, optimizing the agent involves the optimization of thousands of parameters, as is the case of NN controlled agents. Based on that idea, a number of optimization algorithms have been proposed for RL, some examples being the following, Q-learning \cite{watkins1992q}, SARSA \cite{rummery1994line}, Deep Q-learning \cite{mnih2013playing} or Deep Deterministic Policy Gradient \cite{lillicrap2015continuous}.
In order to optimize the parameters of NNs, numerical algorithms are one of the most common choices. However, they usually tend to be computationally very expensive. As an alternative, Evolutionary Algorithms (EAs) are used, and among these, Genetic Algorithms (GAs) are the most frequent \cite{yao1999evolving, liu2004evolving, sedki2009evolving, irani2011evolving, NEAT, wieland1991evolving}.

% EAk
Evolutionary Algorithms are natural evolution inspired techniques. As occurs in nature, a set of solutions or individuals, also called a population, is evolved by means of selection, crossover and mutation operations and, in this way, it tries to find an optimal solution. Among the advantages of EAs, they are applicable in a wide range of problems, they do not make any assumptions about the search space and they can be easily adapted to any combinatorial or continuous problems. 

% GAk
In GAs, solutions or individuals of the population are usually encoded as vectors, for example, binary strings or real-coded vectors. Subsequently, every solution is evaluated according to a fitness function. Finally, some random operators are applied to them in order to create new solutions.

% EDA eta EAk
% ZAHARRA: In this paper, we propose using Estimation of Distibution Algorithms \cite{larraEDA} (EDAs) to optimize the inner parameters of NNs in RL tasks. EDAs belong to the field of Evolutionary Algorithms (EAs). EAs are nature inspired algorithms that starting with a randomly generated population of solutions perform selection, crossover and mutation operations to evolve population's individuals and try this way to find the optimum or at least good solution.   

In this paper, we propose using another paradigm among EAs, an Estimation of Distribution Algorithm \cite{larraEDA} (EDA) in order to optimize the inner parameters of NNs in RL tasks. EDAs, like GAs, also start with a randomly generated population of solutions. Each solution is evaluated according to a fitness function, and the most promising ones are selected following a predefined procedure. Next, a probability distribution is learned from a set of selected individuals. Afterwards, the new generation is composed of survivors and new individuals generated by sampling the learned probability distribution from the previous step. The process is repeated until a maximum fitness value, maximum number of generations or convergence is reached.

% Larra
In other works \cite{larraEDA}, some EDAs are compared with the Stochastic Gradient Descent (SGD) technique, which is the usual choice to pair with the popular backpropagation method in various supervised learning problems. In the conducted experiments, EDAs proved that they have a competitive performance despite being outperformed by SGD.

There are various types of EDAs depending of the kind of the probabilistic model they learn. In this paper, the Univariate Marginal Distribution Algorithm (UMDA$_c$, \textit{c} stands for continuous domain) is used \cite{larraEDA}.
% Evaluation GA vs UMDA$_c$
In order to prove the validity of using UMDA$_c$ for evolving NNs (NN-UMDA$_c$) in RL tasks, the OpenAI Gym \cite{openAIgym} toolkit was adopted. This toolkit is designed to train and test RL algorithms in many different tasks, such as walking, balancing or playing Atari games.
Particularly, we compared the performance of NN-UMDA$_c$ with NN-GA in four different problems from OpenAI's Gym. Experimental results showed a superior performance when using UMDA$_c$.

% Paper structure
The remainder of the paper is organized as follows: in the next
section, some background notes in NN and UMDA$_c$ are given. Afterwards, in section 3, RL problems are described. Later, in section 4, optimization of NN agents by UMDA$_c$ is explained. In section 5, experimental design, obtained measures and a statistical analysis on the results are presented. Finally, in section 6, conclusions and future challenges are given. 

\section{Background}
In this section, we introduce the necessary background on NNs and UMDA$_c$, giving a detailed explanation on how they work. Later, the background from this section is used for explaining the algorithm proposed in this paper.

\subsection{Neural Networks}

Neural Networks are biologically inspired computational models. They were first proposed in 1943 by Warren McCullough and Walter Pitts \cite{mcculloch1943logical}. However, it was not until the 80's that huge advances in computer science finally enabled the computation of complex NNs. Hence, the capability of NNs to solve problems was boosted \cite{wang2017origin}.

NNs are networks of simple processing units called perceptrons, simple computational models of biologic neurons. Perceptrons are given input data in the form of an $n$ size vector. Then, each element of the vector is multiplied by a real number given by the weights of the neuron and all these values are summed together. Usually, a bias is added to the input of the perceptron in order to shift the perceptron's function. 
% Bias is usually given a value of 1, and as every input element, it is multiplied with its respective weight. 
After multiplying the input vector with weights, the results are summed together and an activation function is applied. The activation function gives non-linearity to the model and it can also be used to normalize the output value to a certain range. To give an example, the sigmoid function ($\sigma$) is one of the most widely used activation functions. Also, \textit{tanh} activation function is known to normalize the output vector's values from -1 to 1. Therefore, the output of a perceptron can be described as
\[
    y_{i} = \sigma \bigg( \sum_{j=1}^{n} w_{ij} x_{j} + b_{i} \bigg)
\]

\begin{figure}[h]
    \centering
    \includegraphics[width=0.25\textwidth]{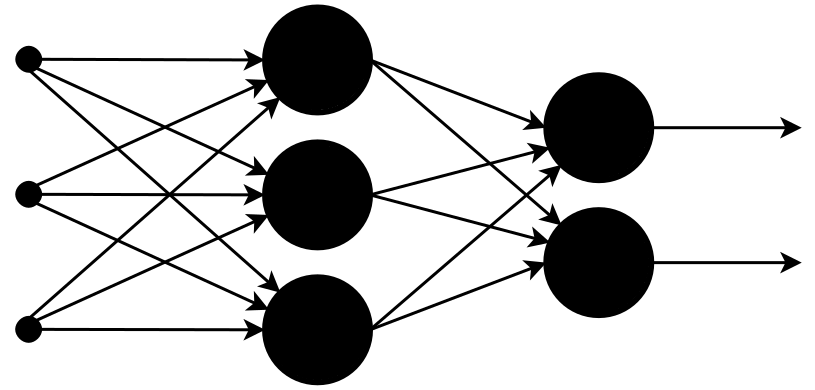}
    \caption{Feed-forward and single hidden layer NN.}
    \label{fig:feed-forward}
\end{figure}

where the output of the perceptron $i$ is $y_{i}$, $x_{j}$ stands for $j$th element of the input vector, $b_{i}$ is the bias of the perceptron and $\sigma$ is the activation function.

When the output of a perceptron turns into the input of another, perceptrons are connected. In NNs, perceptrons are connected to others in the form of layers. This gives  NNs the capability to approximate any kind of function \cite{cybenko1989approximation, hornik1991approximation}. Commonly, a NN is composed of an input layer, some intermediate layers, known as hidden layers, and a final output layer. Depending on how layers are connected, different type of NNs are created. 
% There are three main types of NNs, feed-forward (Figure \ref{fig:feed-forward}), recurrent (RNN) \cite{mikolov2010recurrent} and convolutional (CNN) \cite{krizhevsky2012imagenet}.
Due to the their simplicity, densely connected feed-forward NNs (FNNs) are used in this paper.
In densely connected FNNs, every perceptron of a layer, $h_{i}$, is connected to every perceptron in the next layer, $h_{i+1}$, and there are no loops or cycles between perceptrons, Figure \ref{fig:feed-forward}.
%In this paper, densely connected feed-forward NNs (FNN)  are used, due to their simplicity. Every perceptron of a layer $h_{i}$ is connected to every perceptron in the next layer $h_{i+1}$ (see Figure \ref{fig:feed-forward}). 

%In RNNs, on the contrary of feed-forward NNs, data can flow backwards, and thus creating a internal state or memory. They are vastly used in cases were the input data is a sequence, as is the case of natural lenguage processing \textcolor{red}{[?]}. CNNs consist of various convolutional and pooling layers, due to the fact that they require less computational effort to deal with, they are used for imagery tasks. Frecuently, different kinds of NNs are combined, as is the case of many tasks were images are involved, combining CNN with feed-forward NNs \cite{krizhevsky2012imagenet,lawrence1997face,maturana2015voxnet,rastegari2016xnor,ciresan2011flexible}.

Training NNs is the process where inner parameters of NNs (weights) are optimized in order to minimize the error of the output with respect to the desired output or \textit{target}. This error value is often called the \textit{loss}, and is calculated with a loss function. One such option is the mean squared error or \textit{MSE}, and it is defined as follows, 
\begin{equation*}
    MSE = \frac{1}{n} \sum_{i=1}^{n} (t_{i} - y_{i})^2
\end{equation*}
where $n$ is the number elements in the output vector, $y_{i}$ is the $i$th output of the NN and $t_{i}$ is the expected or target value of $y_{i}$. The computed loss is then propagated backwards in a process called back-propagation, where the parameters of the NN are optimized. 
% For back-propagation different optimization algorithms exist: stochastic gradient descent \cite{bottou2010large}, RSMprop \cite{tieleman2012lecture}, adam \cite{kingma2014adam}, adadelta \cite{zeiler2012adadelta}...  However, in this paper back-propagation is not used, nor none of the optimization techniques mentioned before. 
Despite the existence of several optimization algorithms to combine with back-propagation \cite{bottou2010large, tieleman2012lecture, kingma2014adam, zeiler2012adadelta}, this step is very time consuming, and thus, we decided to substitute it by using UMDA$_c$.
The optimization algorithm employed to improve the performance of the model is explained in the following sections.

\subsection{Univariate Marginal Distribution Algorithm}

Optimizing the weights of NNs can be seen as a continuous optimization problem, which can be approached by means of Estimation of Distribution Algorithms (EDAs). More precisely, Univariate Marginal Distribution Algorithm for continuous domains, UMDA$_c$, is adopted in this work. UMDA$_c$ was introduced by Larrañaga et al. in 1999, \cite{larraEDA}.

\begin{figure}[h]
    \centering
    \includegraphics[width=0.4\textwidth]{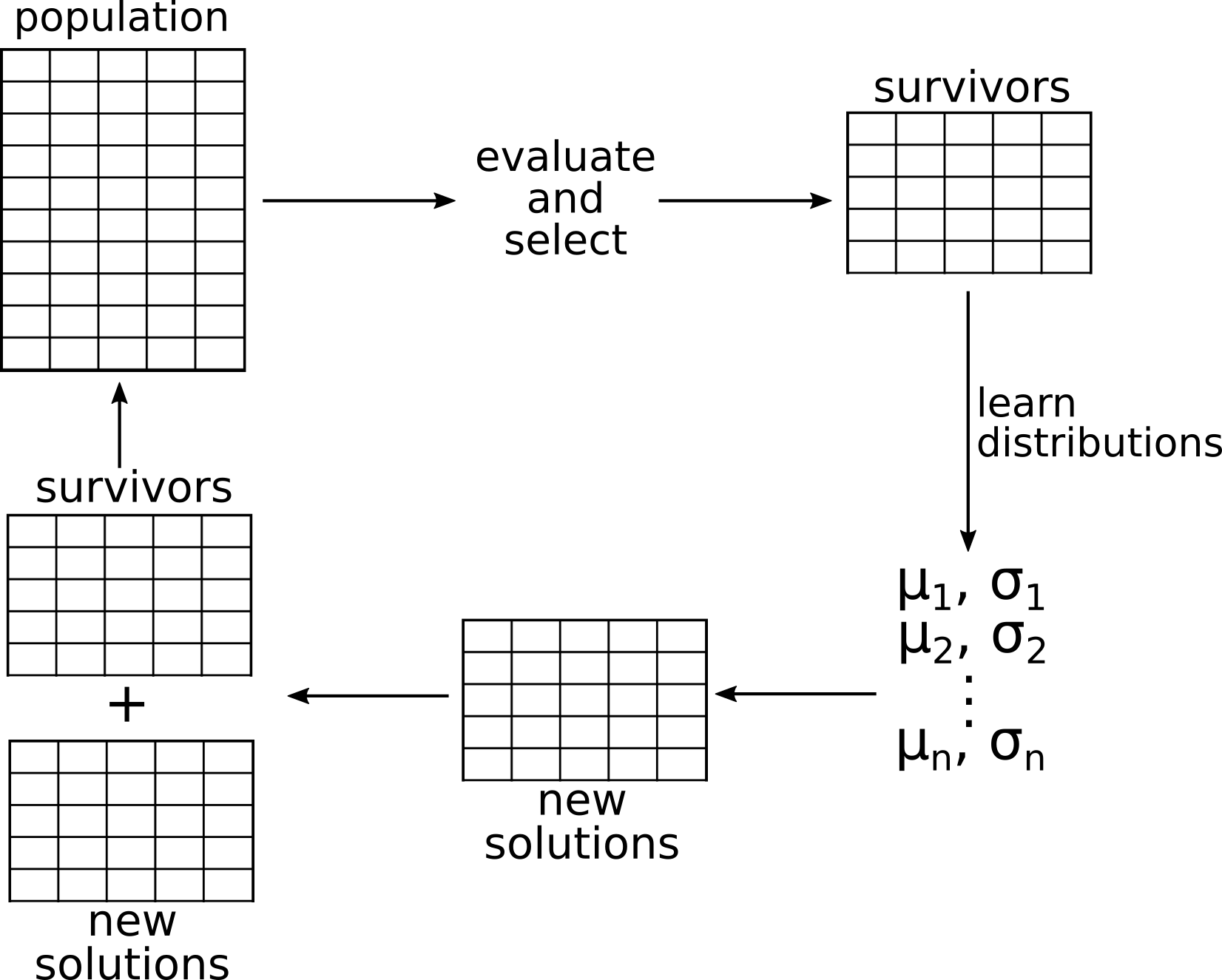}
    \caption{Scheme of the UMDA$_c$ considered for learning the parameters of NNs.}
    \label{fig:umdac_img}
\end{figure}

As with most EDAs, the start point for UMDA$_c$ is a randomly generated population of solutions codified as real-valued vectors. Next, a problem specific evaluation is conducted, thus giving a \textit{fitness} value to every solution of the population. According to the specified \textit{survivor rate}, a set with the best solutions is selected (Figure \ref{fig:umdac_img}). Later, a normal distribution is estimated from the set of survivor individuals on each of the positions in the real valued vectors. In order to create a new set of solutions, random samples from the normal distributions are taken. Finally, the new set of solutions along the solutions from the survivor set will form a new generation or population. The described process is repeated until some convergence criteria are met.

\begin{figure}[h]
    \centering
    \includegraphics[width=0.3\textwidth]{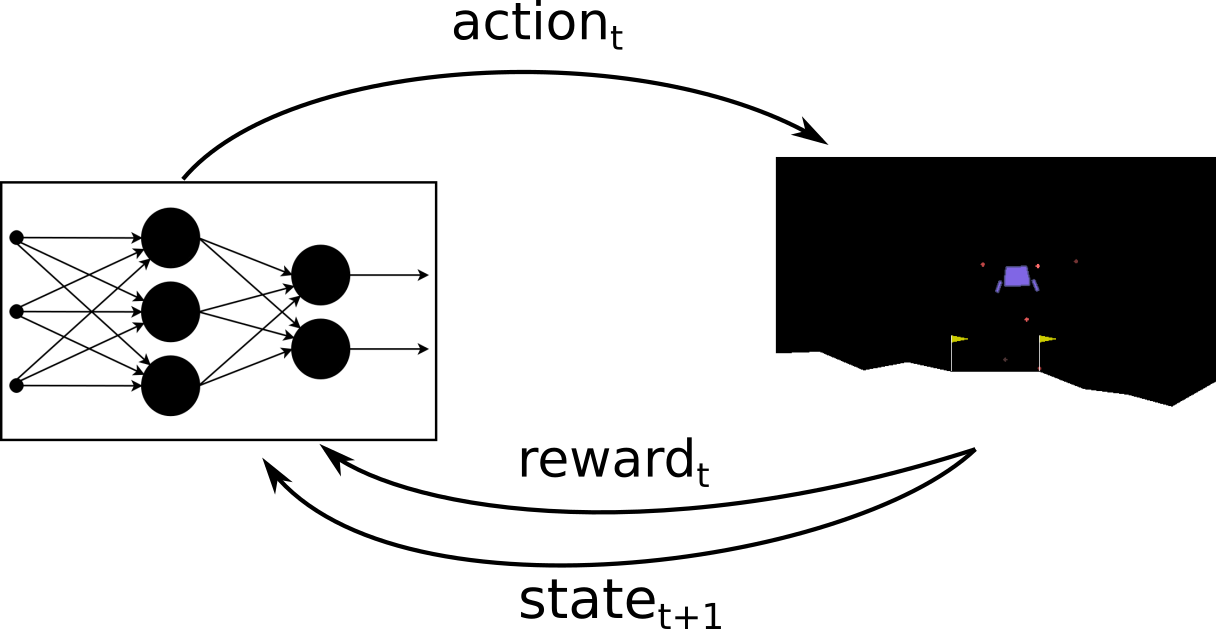}
    \caption{Reinforcement Learning scheme.}
    \label{fig:RL}
\end{figure}

\section{Reinforcement learning}

% In this paper, the purposed algorithm and a GA are evaluated in some RL benchmark problems from OpenAI Gym toolkit \cite{openAIgym}. As an example, \textit{LunarLander-v2} environment is used in this section. In \textit{LunarLander-v2} the agent's (lander) objective is to land between the two flags using the fewest fuel as possible.

As mentioned in the introduction, in Reinforcement Learning (RL) problems the agent is given an observation of the current state of the environment. In the case of OpenAI's Gym environments \cite{openAIgym}, for every frame of the game, the environment returns a \textit{state} vector, $s_{t}$.
In response, the agent takes an action $a_{t}$, and the environment returns a reward value $r_{t}$, and a new state of the game $s_{t+1}$. The goal is to learn an optimum policy that maximizes the cumulative reward value. This interaction of the agent and environment is illustrated in Figure \ref{fig:RL}. 
%For every frame of the game, the environment returns a \textit{state} vector $s_{t}$, Figure \ref{fig:RL}. This vector contains information of coordinates of the lander, distance to ground... 

In this paper, agents are controlled by NNs, and state vectors are the input of NNs. The output of NNs, $out_{n}$, is an $n$ sized vector, $n$ being the number of actions the agent can take. 
In the particular case of \textit{LunarLander-v2} environment \cite{openAIgym} (in which the objective is to land between two flags), $n$ equals 4, as there are four available actions: do nothing, fire left engine, fire main engine and fire right orientation engine.
The index of the highest value element in $out_{n}$ is chosen as the action, $a_ {t}$, to take. As a result, the action is defined as:
\begin{equation*}
    a_{t} = \mathrm{argmax}(out_{n})    
\end{equation*}

After selecting the action $a_{t}$ for the current state $s_{t}$, the action is taken and the environment returns a new state of the environment $s_{t+1}$, and a reward value $r_{t}$. The reward value represents how good or bad the taken action $a_{t}$ was for state $s_{t}$. This process is repeated until the agent dies or the game finishes. Every reward the agent is given is summed as the \textit{total reward}, so after the game finishes is used to evaluate the behavior of the individual, often called \textit{fitness}. 

% Generalization / Stochastic envs
%In order to encourage the model to generalize, the environments are stochastic. For example, in the case of \textit{LunarLander-v2}, the agent starts in the top of the screen with a random inclination and the terrain is randomly generated in every trial. This prevents the model for memorizing a sequence of correct actions, as every game it plays, the optimal sequence of actions to take change. 

\begin{figure*}[b]
    \centering
    \makebox[\textwidth]{\includegraphics[width=0.7\paperwidth]{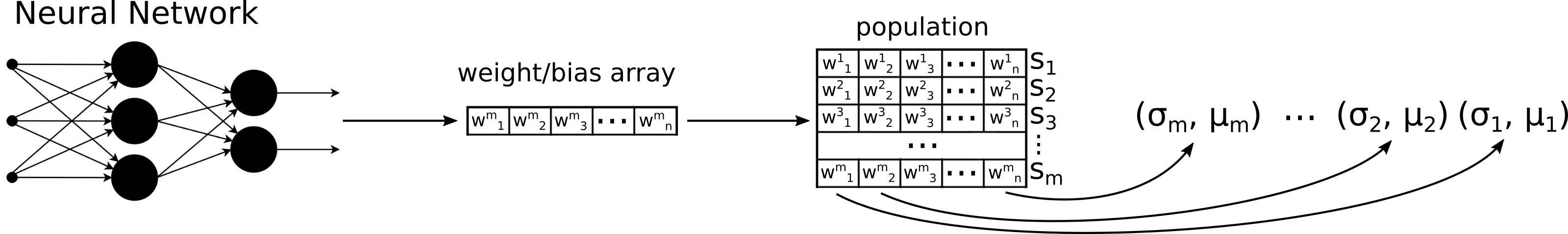}}
    \caption{NN parameter treatment in order to apply UMDA$_c$.}
    \label{fig:Representation}
\end{figure*}

\section{Optimizing NN Agents with UMDA$_c$}
In this section, the contribution of the paper, the optimization of NNs via UMDA$_c$ is presented, henceforth known as NN-UMDA$_c$.

% The aim of this algorithm is to maintain the adaptability to different problems and efficiency of EDAs finding global optimums while conserving the great generalization ability of NNs. In order to reach this goal, UMDA$_c$ is used as a more appropriate approach than GA in order to evolve NNs.

Optimizing the parameters of NNs can be seen as a continuous optimization problem. For this reason, UMDA$_c$ is used as a more appropriate approach than GA in order to evolve NNs, as UMDA$_c$ manages large search spaces more efficiently than GA does. 
Conversely to GAs, UMDA$_c$ is not limited to crossover and mutation, new solutions are sampled from probability distributions, these being more suitable for very difficult search space problems, such as the optimization of the parameters of NNs \cite{larraEDA}. 

% Representation
As aforementioned, in these RL problems we propose controlling the agents by NNs. So, the population of solutions is composed of real-valued vectors representing the parameters of NNs. As a result, the population is a matrix containing vectors of NNs of $n$ dimensions, $n$ being the number of inner parameters of the selected network architecture (see Figure \ref{fig:Representation}). 
Since all the individuals of the population describe the parameters of a unique network architecture, every row of the population matrix has the same length. Each column in the matrix represents a single weight or bias of the selected NN architecture.

Once the population of solutions is evaluated, the ones with the highest fitness values are selected, thus creating a matrix of individuals called survivors, as shown in Figure \ref{fig:umdac_img}. 
% As explained, every column of the matrix corresponds to a specific weight or bias in the network architecture, so for every column of the survivor matrix the mean and variance are calculated in order to create a normal distribution for every weight position. 
% Creation of new solutions
The next step is to create a set of new solutions or individuals. For every column in the survivors matrix, the mean and variance of the column is calculated in order to estimate a normal distribution for a certain parameter of the NN. Then, new values are sampled from the normal distribution, creating new weight values for that specific parameter.

% For every new individual and each weight or bias in the NN architecture, new weight or bias value is sampled from its respective Gaussian distribution, Figure \ref{fig:umdac_img}.

After generating new individuals, individuals that are not in the survivor set are replaced with new ones, thus creating a new population of solutions, Figure \ref{fig:umdac_img}. As the process described is repeated, random samples of the normal distributions learned from the best individuals of each generation will tend to better values in order to maximize the fitness value. 
% Muatation rate eta EDAk
In GAs, a \textit{mutation rate} is commonly used to prevent convergences in local optimums and to add diversity to the population of solutions in order to explore other areas from the solution landscape.
On the contrary, in UMDA$_c$ there is no need for such techniques. As diversification implicitly happens when new solutions are randomly sampled from normal distributions.

% Wight position's importance
%Finally, in NNs a certain weight value has only sense in its specific location, while in another position it could cause a great decrease in the NN's performance. For this reason, a normal distribution is calculated for every weight $w_ {ij}$ architecture with the $w_{ij}$ weight from every selected solution of the population, as the same probability distribution could not have sense for other weight location.

%\begin{figure}[t]
%    \centering
%    \makebox[\textwidth]{\includegraphics[width=0.5\paperwidth]{img/representation.png}}
%    % \includegraphics[width= 0.3pt \paperwidht]{img/representation.png}
%    \caption{NN parameter treatment in order to apply UMDA$_c$.}
%    \label{fig:Representation}
%\end{figure}

\section{Experiments}

In order to validate our proposal, NN-UMDA$_c$ and NN-GA were evaluated in four RL tasks. More specifically, \textit{CartPole-v0}, \textit{LunarLan\-der-v2}, \textit{LunarLanderCountinous-v2} and \textit{Bipedal-Walker-v2} environments from  OpenAI Gym toolkit \cite{openAIgym} were adopted for evaluation. 

\subsection{Experimental Setup}

To define the scale of a problem, the number of trainable parameters in the selected NN architecture was taken, $n$. 
For the proposed problems in these experiments, the simplest NN architecture for each task was chosen. Consequently, the proposed architectures contain an input layer with the size of the state vector the environment returns, and an output layer with the size of the number of available movements. 
Some parameters of both algorithms are under $n$, as those parameters depend on the scale of the task to be solved \footnote{The parameters were set by tuning their values manually and carrying out non-exhaustive experiments.}. NN-UMDA$_c$ and NN-GA share the same parameters, in order to be evaluated under the same conditions. With the exception of \textit{mutation rate} in the GA\footnote{
The GA used in this paper follows the same steps as UMDA$_c$, besides the sampling of new solutions. The GA, instead of sampling new solutions from normal distributions, samples the values from a uniform distribution from the survivors set of solutions, applying some random noise to the samples taken, defined by the \textit{mutation rate}.
}, the rest of the parameters are equally set in both algorithms in order to evaluate under the same conditions. In the presented experiments, the \textit{mutation rate} has a constant value of 0.1. The rest of the parameters for both algorithms are:

\begin{itemize}
    \item Population size: $6n$
    \item Number of generations: $3n$
    \item Number of individual evaluations: $3$
    \item Survivor rate: $0.5$
\end{itemize}

The population size depends on the dimension of the problem. For larger problems, in which the solution space is much larger and finding an optimum solution is harder, a higher number of solutions per generation is needed to solve the problem. The same occurs with the number of iterations or generations needed.

% Generalization / Stochastic envs
In order to encourage the model to generalize, the environments are stochastic. 
% For example, in the case of \textit{LunarLander-v2}, the agent starts in the top of the screen with a random inclination and the terrain is randomly generated in every trial.
This prevents the model from memorizing a sequence of correct actions, as for every game it plays, the optimal sequence of actions to take changes. 
% Since environments are stochastic, solutions that performed correctly in a trial may not perform equally in other trials.
Consequently,  solutions that performed correctly in a trial may not perform equally in other trials. Therefore, to have an overall understanding of how well an individual performs, the same solution is evaluated 3 times, and the average of total rewards is taken as the fitness value of the given individual. 
The number of times each solution is evaluated is referred to as the \textit{number of individual evaluations}. The number of solutions to select as survivors is defined as the \textit{survivor rate}. In these experiments, the \textit{survivor rate} is always $0.5$, meaning a half of the population is going to remain unchanged while the other half can be replaced with new solutions every generation.

Presented experiments were carried out on a machine of the following characteristics: 8 cores Intel Xeon Skylake processor and 7.2GB of RAM memory. 

\subsection{Performance evaluation}
\begin{figure*}
    \centering
    \includegraphics[width=0.8\textwidth]{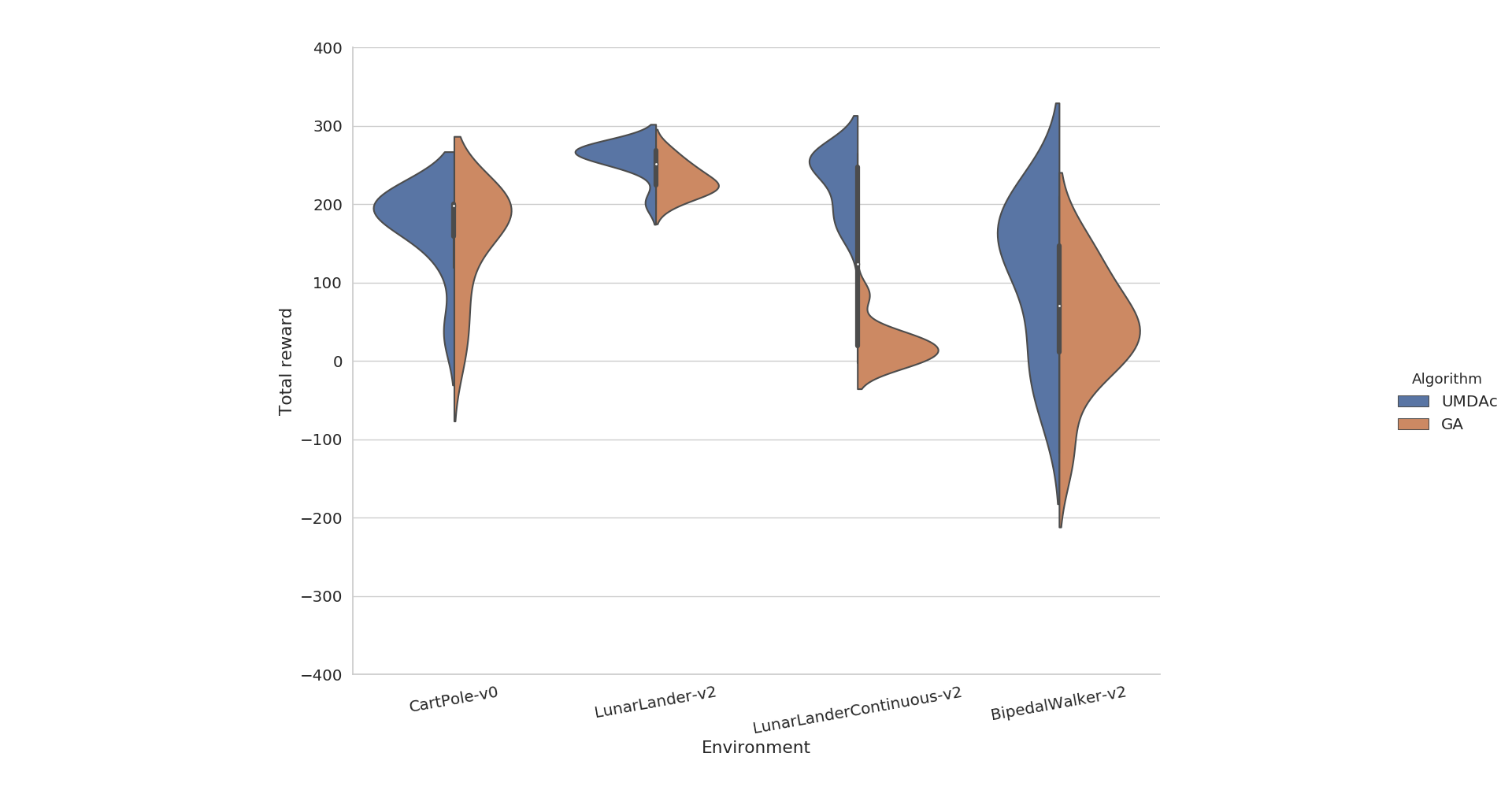}
    \caption{The performance of both algorithms in the given environments.}
    \label{fig:Experiment}
\end{figure*}

% ZAHARRA
% In order to collect data for later assessment, both algorithms were run 10 times on each benchmark problem. Furthermore, the best solution of each run was evaluated 100 times. So, 1000 evaluation samples from each algorithm were collected for every problem. The results obtained are summarized in Table \ref{tab:averages} and Figure \ref{fig:Experiment}. In Table \ref{tab:averages}, the averages of the obtained \textit{total rewards} are given, while in Figure \ref{fig:Experiment}, the results are plotted as violin plots, representing the kernel density estimations of the \textit{total-rewards} obtained when evaluating the best results of both algorithms.
% EGUNERATUA
In order to collect data for later assessment, both algorithms were run 10 times on each benchmark problem. Furthermore, the best solution of each run was evaluated 100 times, and the \textit{average total reward} was calculated. The results obtained are summarized in Table \ref{tab:averages} and Figure \ref{fig:Experiment}. In Table \ref{tab:averages}, the averages of the obtained \textit{total rewards} are given, while in Figure \ref{fig:Experiment}, the results are plotted as violin plots, representing the kernel density estimations of the \textit{average total rewards} obtained when evaluating the best results of both algorithms.

\begin{table*}
\centering
\begin{tabular}{l|rrrrr}
Algorithm & CartPole-v0 & LunarLander-v2 & LunarLanderContinuous-v2 & BipedalWalker-v2 \\\hline
NN-UMDA$_c$ & 173.26 & 258.80 & 229.34 & 108.80\\
NN-GA & 155.60 & 231.30 & 21.22 & 39.71
\end{tabular}
\caption{\label{tab:averages}Average total rewards of the algorithms in every environment.}
\end{table*}

\subsubsection{CartPole-v0}

Firstly, algorithms were tested in the \textit{CartPole-v0} environment \cite{openAIgym}. In this classic control task, a pole is joined to the cart by an non-actuated joint, and the objective is to maintain the pole in balance. There are only two available actions: push the cart to the left or push it to the right. In \textit{CartPole-v0}, the state is a four element vector, representing: cart position, cart velocity, pole angle and pole velocity at tip. The game is considered solved after an average reward of 195.0 over 100 consecutive trials. In this task, according to the chosen NN architecture, the scale of the problem, $n$, is 10. 

% Even though both algorithms were able to solve the game in almost every trial, NN-GA failed to solve the game considerably more times than NN-UMDA$_c$ did. As observed in Figure \ref{fig:Experiment} the scores of NN-UMDA$_c$ in almost every trial were 200. However, NN-GA did not generalize so well, and in some cases the obtained scores dropped below 100.
NN-UMDA$_c$ and NN-GA were able to solve the game in almost every trial.
The \textit{total rewards} of NN-UMDA$_c$ were almost always 200. However, NN-GA did not generalize so well, and in some cases the obtained \textit{total rewards} dropped below 100.

\subsubsection{LunarLander-v2}

The next step was to evaluate both algorithms in a more difficult task. \textit{LunarLander}'s state vector size and the number of available actions are twice those of \textit{CartPole-v0}. Consequently, the complexity increased considerably, having $n=36$ for this game. As previously mentioned in the RL section, in the \textit{LunarLander-v2} environment the objective of the agent is to land between the two flags. The game is solved if an average reward of 200 or more is achieved in 100 runs.

As shown in Figure \ref{fig:Experiment}, both algorithms solved the proposed task and performed similarly. However, NN-UMDA$_c$ was able to obtain slightly higher scores than NN-GA. 

\subsubsection{LunarLanderContinous-v2} 

To test the game in a continuous control task, the \textit{LunarLanderContinous-v2} problem was chosen. Both \textit{LunarLander} environments are equal except for having an action vector of two real values from -1 to +1 for this environment. This real number vector controls the thrust of the main and both side engines.

From the last three presented environments, \textit{Lunar\-Lan\-der\-Con\-ti\-nous-v2} has the greatest difference in performance between both algorithms. The results obtained, Figure \ref{fig:Experiment}, showed that NN-UMDA$_c$ solved the game in most of the trials, even above the score considered to solve the game (200), while NN-GA was not able to solve the game in most of the cases, resulting in a much poor performance.

\subsubsection{BipedalWalker-v2} 

The results of the last experiment motivated the comparison of the algorithms in a task of larger search space. In order to test both algorithms in a more challenging benchmark, \textit{BipedalWalker-v2} was adopted. In this continuous control task, in order to maximize its reward, the bipedal agent has to move forward as fast as possible without falling. The size of the state vector increases considerably compared to the other presented environments, as $n=100$. 
For this specific problem, due to computational limitations, parameters under $n$ were tuned in order to reduce the training time. The parameters changed for \textit{BipedalWalker-v2} are: 

\begin{itemize}
    \item Population size: $3n$
    \item Number of generations: $1n$
\end{itemize}
As a result, neither NN-UMDA$_c$ nor NN-GA were able to solve the game (rewards above 300). However, the goal was not to solve the game, but to get useful data to compare the performance and behavior of both algorithms. As expected, NN-UMDA$_c$ obtained considerably higher rewards than NN-GA in most of the evaluations conducted. 

% \subsection{Time consumption}

\subsection{Statistical analysis}
In order to statistically assess the performance results obtained from the compared algorithms, we have followed the Bayesian approach presented in~\cite{benavoli2017}, as it can provide a deeper insight into the results than the classical null hypothesis significance tests. Particularly, we have used the Bayesian equivalent of the Wilcoxon's test\footnote{We have used the implementation available in the development version of the \textbf{scmamp} R package~\cite{calvo2016} available at \url{https://github.com/b0rxa/scmamp}.}. Conversely to the usual null hypothesis statistical analysis, Bayesian methods permit to answer explicitly questions such as, {\it which is the probability that a given algorithm is the best one?} Not only that but, due to the nature of the Bayesian statistics, it naturally provides us with information about the uncertainty about the behavior of the algorithms after the data from the experimentation has been introduced.

In this particular case, the Bayesian analysis was conducted on the average reward obtained by each algorithm in the 10 repetitions. Under this analysis the goal is to determine how likely will be NN-UMDA$_c$ to beat NN-GA when run on each of the benchmarks considered. To that end, the procedure used requires the definition of what is understood as `practical equivalence' or 'rope' in~\cite{benavoli2017}, i.e. the performance difference below which both algorithms can be considered with same performance. In our case, we decided that both approaches have equivalent performance when the difference in reward is smaller than $0.1$. In the particular case of this experimentation, the Bayesian analysis permits to explicitly calculate the posterior probability distribution on the probability of each scenario being the most likely i.e. NN-UMDA$_c$, NN-GA or {\it Rope}. To that end, as in any Bayesian model, a prior knowledge needs to assumed about the probability of each of the algorithms being the best. In this case, we assume uniform probability on all the possible probability distributions of the three options. Then, the results obtained from the experimentation are incorporated in order to update our belief accordingly, calculating this way the posterior distribution\footnote{As this paper is not devoted on Bayesian analysis, we refer the interested reader to address the papers~\cite{benavoli2017,Calvo2016} for further details.}.

\begin{figure}
        \centering
        \subfigure[{\small {\it CartPole}. UMDA$_c$: 0.497, GA: 0.372, Rope: 0.129}]{\label{fig:cartpole} \includegraphics[width=0.475\textwidth]{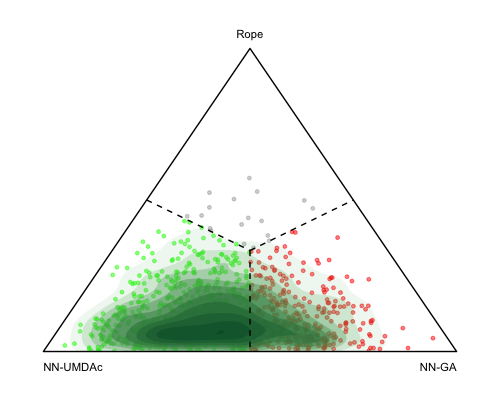}}\hspace{0.5cm}
        \subfigure[{\small {\it LunarLander}. UMDA$_c$: 0.952, GA: 0.040, Rope: 0.006}]{\label{fig:lunar} \includegraphics[width=0.475\textwidth]{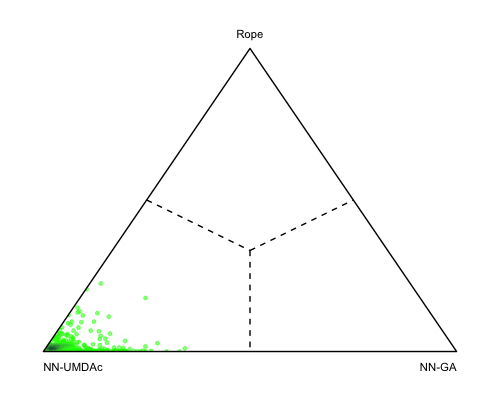}}
        \subfigure[{\small {\it LunarLanderContinuous}. UMDA$_c$: 0.993, GA: 0.00, Rope: 0.006}]{\label{fig:lunarcont} \includegraphics[width=0.475\textwidth]{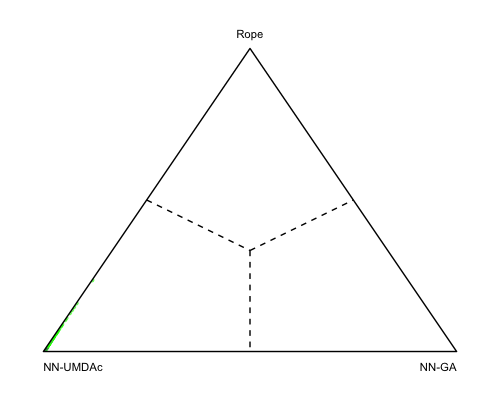}}\hspace{0.5cm}
        \subfigure[{\small {\it BipedalWalker}. UMDA$_c$: 0.821, GA: 0.005,  Rope: 0.172}]{\label{fig:bipedal} \includegraphics[width=0.475\textwidth]{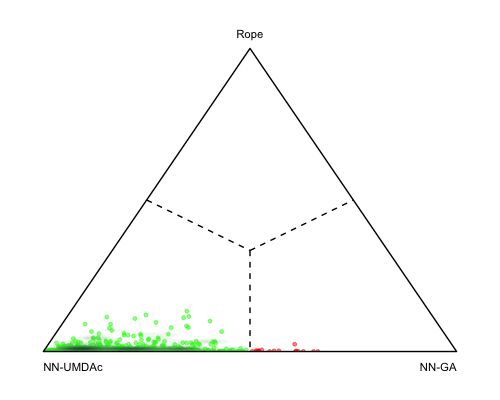}}
    \caption {\small Simplex plots of the results obtained in the four benchmark environments, and the expected posterior probabilities of each algorithm being the best algorithm.} 
    \label{fig:simplex}
\end{figure}

Results of the analysis are depicted in Fig. \ref{fig:simplex} as simplex plots. These plots are triangle diagrams where the vertices represent the probability vectors $\{(1,0,0),(0,1,0),(0,0,1)\}$. The points in the plot represent a sampling of the posterior distribution of the probability of win-lose-tie. The closer a point is to the NN-UMDA$_c$ vertex of the triangle (or, equivalently, to NN-GA or {\it Rope} vertices), the more probable it is for NN-UMDA$_c$ to have better performance (or equivalently, NN-GA or both algorithms being equal). Therefore, the three areas delimited by the dashed lines show the dominance regions, i.e., the area where the highest probability corresponds to its vertex.

Fig.~\ref{fig:simplex} shows that depending on the environment optimized, very different scenarios can be observed. As regards {\it LunarLander} and {\it LunarLanderContinuous}, the probability mass of the posterior is closed to the vertex NN-UMDA$_c$ (see Figs.~\ref{fig:lunar} and~\ref{fig:lunarcont}). This fact is confirmed by the average posterior probabilities for each situation (NN-UMDA$_C$ being better, equal or worse than NN-GA), which points out that the expected probability of NN-UMDA$_C$ being better is 0.92 and 0.993, respectively. We also see that the spread of the points in the two plots is very low, which indicates that the variance of the posterior distribution is small, i.e. there is almost no uncertainty about the result of analysis: NN-UMDA$_C$ performs much better than NN-GA.

With respect to the results in {\it BipedalWalker} environment, we see that most of the points are in the region of NN-UMDA$_c$ indicating that this algorithm has larger probability to show better performance. This is corroborated with the expected probability $0.821$. Nevertheless, in this case, the posterior variance is larger than previously, and suggests that,  occasionally,  NN-GA is preferred to NN-UMDA$_c$.

Finally, in {\it CartPole}  we see that the expected posterior probability of the EDA approach is slightly higher (see Fig.~\ref{fig:cartpole}). However, we see that, on the one hand, the points in the simplex plot are very sparse. This shows that the variance of the posterior distribution is high (large uncertainty), and more executions are needed to draw solid conclusions. On the other hand, we also see that the points are spread in the regions relative to UMDA$_c$ and GA, and a few in the {\it Rope}. This fact shows that the uncertainty related to which algorithm is better is high, i.e., with the executions at hand, we can say it looks that NN-UMDA$_C$ and NN-GA have similar performance being the first slightly better.

Finally, as the practical equivalence threshold was set very small when compared to the obtained rewards, almost never both algorithms are considered equivalent during the analysis. Therefore, the expected posterior probability of the {\it Rope} is very low in all the considered environments.

\section{Conclusions and future challenges}

% In this paper, we have introduced an alternative to GAs in order to evolve NNs. We have demostrated that the proposed algorithm was capable of resolving every given tasks more efficiently than the most commonly used algorithm (GA) to evolve NNs. The presented algorithm was able to resolve the given task more reliably than the GA, and in some cases the obtained score was much greater than the needed to resolve the problem. What is more, our algorithm performed as great as many other algorithms tested in the same benchmarks.

In this paper, we introduced an alternative to GAs in order to adjust the parameters of NNs for RL tasks. We also proved that the proposed algorithm, NN-UMDA$_c$, was capable of solving every given task more efficiently than NN-GA, a greatly used algorithm to evolve NNs. NN-UMDA$_c$ was not only able to resolve the given task more reliably than NN-GA, but in some cases the obtained score was greater than the considered threshold to solve the problem.  
% \textcolor{red}{What is more, NN-UMDA$_c$ performed as great as many other algorithms tested in the same benchmarks.}

The experiments in this paper were conducted to compare the performance of NN-GA with respect to NN-UMDA$_c$, and obtain some empirical data for the evaluation of both algorithms. However, further research has to be done in order to test the presented algorithm in more complex and advanced scenarios. In the following, we enumerate three possible lines for future research.

\begin{enumerate}
    
    \item In the presented experiments, NNs were always feed-forward single layer NNs. As the proposed algorithm can be applied to many types of NNs with very little effort, more advanced NN architectures could be used in order to face harder problems. For example, CNNs in scenarios where the input data are images, or RNNs to deal with time dependant actions.
    
    \item EAs are great at efficiently approaching near global optimum solutions. After evolving a NN, this could be used to pre-train another NN. In this process, the knowledge of the near optimal solution would be transferred to the other NN. Finally, the pre-trained NN could be optimized (using another algorithm such as DQL \cite{mnih2013playing}) to fine-tune towards the global optimum.

    % \item Cascade optimization of neural network architecture. Starting from small neural network architectures (few neurons and layers), gradually increase the size of neural networks of all individual by adding some neurons or layers while training.
    
    \item % in other works \cite{WorldModelsDavidHa, higgins2017darla}, a special type of NN, Variational Autoencoders
    A special type of NN, Autoencoders, could be used to learn a reduced dimensionality representation of the observation space \cite{hinton2006reducing}. Autoencoders are previously trained to learn this simpler representation. Consequently, much simpler models are trained to learn to solve the required task.
    As regards the algorithm presented in this manuscript, populations of much simpler NNs could be trained. Solutions could be fed with a much smaller dimension state vector, thus reducing the training time and computational load considerably.
    
    \item To conclude, in this paper, UMDA$_c$ is only compared with a GA. For better understanding the convenience of optimizing the parameters of NNs with UMDA$_c$, the proposed algorithm should be compared with more types of EAs.
    
\end{enumerate}

\section*{Acknowledgements}
This Work has been partially supported by TIN2016-78365R (Spanish Ministry of Economy, Industry and Competitiveness). We would like to gratefully acknowledge the support of Unai Garciarena in the revision of the manuscript.

%\begin{acks}
%Work has been partially supported by TIN2016-78365R (Spanish Ministry of Economy, Industry and Competitiveness). We %would like to gratefully acknowledge the support of Unai Garciarena in the revision of the manuscript.
%\end{acks}

\bibliographystyle{unsrt}  
\bibliography{references}  %%% Remove comment to use the external .bib file (using bibtex).

\end{document}